%% file: main.tex
\documentclass[10pt,twocolumn,letterpaper]{article}

\usepackage[pagenumbers]{cvpr} 

\usepackage{multirow}
\usepackage{multicol}
\usepackage{soul}
\usepackage[table]{xcolor}
\usepackage[most]{tcolorbox}
\usepackage{subcaption}
\usepackage{stfloats}

\definecolor{cvprblue}{rgb}{0.21,0.49,0.74}
\usepackage[pagebackref,breaklinks,colorlinks,allcolors=cvprblue]{hyperref}

\def\paperID{10940} 
\def\confName{CVPR}
\def\confYear{2026}

\title{Looping Back to Move Forward: Recursive Transformers for Efficient and Flexible Large Multimodal Models}

\author{
Ruihan Xu$^{1}$, Yuting Gao$^{2}$, Lan Wang$^{2}$, Jianing Li$^{2}$, Weihao Chen$^{2}$, Qingpei Guo$^{2}$,\\
Ming Yang$^{2}$, Shiliang Zhang$^{\dagger 1}$\\
$^{1}$
State Key Laboratory of Multimedia Information Processing,\\School of Computer Science, Peking University\quad
$^{2}$AntGroup\\
}

\begin{document}
\maketitle
\input{sec/0_abstract}    
\input{sec/1_intro}

\input{sec/2_related}

\input{sec/3_method}
\input{sec/4_experiments}
\input{sec/5_conclusion}
{
    \small
    \bibliographystyle{ieeenat_fullname}
    \bibliography{main}
}

\input{sec/X_suppl}

\end{document}

%% file: sec/0_abstract.tex
\begin{abstract}


Large Multimodal Models (LMMs) have achieved remarkable success in vision–language tasks, yet their vast parameter counts are often underutilized during both training and inference.
In this work, we embrace the idea of looping back to move forward: reusing model parameters through recursive refinement to extract stronger multimodal representations without increasing model size. We propose RecursiveVLM, a recursive Transformer architecture tailored for LMMs.
Two key innovations enable effective looping: i) a Recursive Connector that aligns features across recursion steps by fusing intermediate-layer hidden states and applying modality-specific projections, respecting the distinct statistical structures of vision and language tokens; ii) a Monotonic Recursion Loss that supervises every step and guarantees performance improves monotonically with recursion depth.
This design transforms recursion into an on-demand refinement mechanism: delivering strong results with few loops on resource-constrained devices and progressively improving outputs when more computation resources are available.
Experiments show consistent gains of +3\% over standard Transformers and +7\% over vanilla recursive baselines, demonstrating that strategic looping is a powerful path toward efficient, deployment-adaptive LMMs.


\end{abstract}

%% file: sec/1_intro.tex
\section{Introduction}
\label{sec:intro}


\begin{figure}[t]
    \centering
    \includegraphics[width=\linewidth]{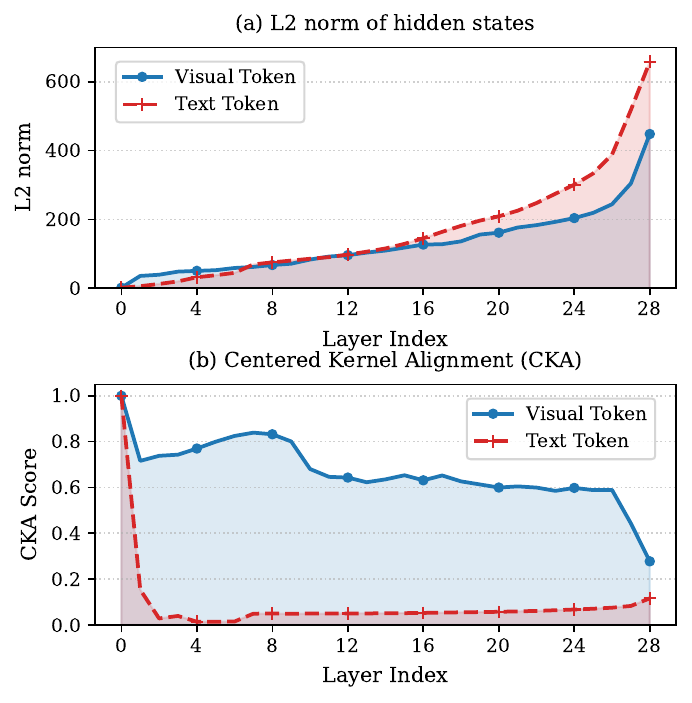}
    \caption{Statistical results in Ming-Lite-Omni~\cite{ming_omni}. (a) The L2 norm of the hidden states in each layer indicates misalignment in feature scale. (b) The Centered Kernel Alignment (CKA) between the hidden states and the input embeddings indicates a feature distribution misalignment in vanilla recursive models.}
    \label{fig:misalignment}
\end{figure}

Large Multimodal Models (LMMs), such as GPT-4V~\cite{gpt4v}, Gemini~\cite{gemini,gemini1_5,gemini2_5}, and Qwen-VL~\cite{qwen,qwen2vl,qwen2_5vl}, have achieved remarkable performance on vision–language tasks like visual question answering and image captioning~\cite{vqa,image_captioning,multimodal}.
Yet this success relies on massive parameter counts that are typically engaged only once during inference, leaving much of their representational capacity underutilized.
Recent work in language modeling demonstrates that reusing the same parameters across multiple iterative passes can significantly enhance reasoning and representation quality without increasing model size~\cite{mobilellm,saunshi2025reasoning,relaxed_recursive_transformers,mor,ouro}.
Recursive Transformers, theoretically capable of simulating programmable computation~\cite{looped_programmable}, have already advanced the state of the art in Large Language Models (LLMs). These findings suggest that recursive refinement offers a promising path toward more efficient LMMs—\textbf{looping back over existing weights to move multimodal understanding forward}.

However, extending recursion from unimodal LLMs to multimodal LMMs is far from straightforward. Unlike homogeneous text sequences, LMMs process heterogeneous vision and language tokens whose representations evolve along divergent trajectories across layers (see Figure~\ref{fig:misalignment}). This intrinsic modality gap introduces unique alignment and stability challenges that invalidate naive recursive designs developed for LLMs.

The first challenge is \textbf{severe feature misalignment between recursion steps}. A vanilla recursive model feeds the output of one step directly as input to the next (see Figure~\ref{fig:taxonomy}(b)), but hidden states in LMMs diverge significantly in both scale and distribution across layers. As shown in Figure~\ref{fig:misalignment}(a), the L2 norm of hidden states grows rapidly with depth, causing scale mismatch that leads to numerical instability when deep features are reused as shallow inputs. Figure~\ref{fig:misalignment}(b) further reveals low Centered Kernel Alignment (CKA)~\cite{cka} between deep-layer features and input embeddings, indicating semantic distribution drift. Compounding this, vision and language tokens exhibit inherently different statistical patterns, making a unified alignment strategy ineffective.

The second challenge lies in the \textbf{lack of flexible, step-wise predictability}. Most existing recursive models are designed to produce a final output only at a pre-specified recursion depth. While recent works introduce token-level dynamic gating~\cite{ouro} or routing mechanisms~\cite{mor} to adjust computation per token, they still optimize for a single target depth and do not ensure consistent quality across intermediate steps. Consequently, early or truncated recursions often yield suboptimal or even degraded predictions, making them unreliable under variable latency or energy constraints. In real-world deployment, however, a model must support on-demand inference: delivering usable results after just one or two steps on edge devices, while progressively refining outputs in cloud environments. This demands not only valid predictions at every step, but also monotonic performance improvement with recursion depth, a property absent in current approaches and essential for trustworthy adaptation.

To address these challenges, we propose RecursiveVLM, the first recursive Transformer architecture tailored for LMMs.
As illustrated in Figure~\ref{fig:taxonomy}, standard LMMs (left) perform a single forward pass and vanilla recursive models (middle) naively feed hidden states back as inputs, suffering from severe misalignment. Differently, our design (right) introduces a dedicated Recursive Connector to bridge recursion steps. This connector resolves cross-step misalignment by (i) normalizing feature scales via RMSNorm, (ii) mapping deep representations back to the input embedding space through a residual MLP trained with continual learning, and (iii) employing modality-specific projections for visual and textual tokens. It further enriches inputs by fusing features from multiple intermediate layers.
Complementing this architectural innovation, we introduce a Monotonic Recursion Loss that supervises every recursive step and explicitly enforces non-decreasing performance with depth, enabling valid predictions at any computational budget. Together, they transform recursion into a deployment-adaptive refinement mechanism, unlocking stronger performance from the same parameters while supporting dynamic resource constraints.

Experiments across 8 standard vision–language benchmarks show that RecursiveVLM, trained with recursive refinement, outperforms the standard non-recursive baseline by 1-2\% even when using only a single recursion step at inference, demonstrating that our architecture learns more effective representations through recursive training alone. At the second recursion step, performance improves by +3\% compared with the non-recursive baseline, confirming monotonic gains and validating our design for deployment-adaptive inference. To the best of our knowledge, this is an original attempt on a recursive Transformer architecture for LMMs. Its success may inspire further efforts on compact LMMs through recursively using parameters.

%% file: sec/2_related.tex
\section{Related Works}
\label{sec:related}


\begin{figure*}
    \centering
    \includegraphics[width=0.9\linewidth]{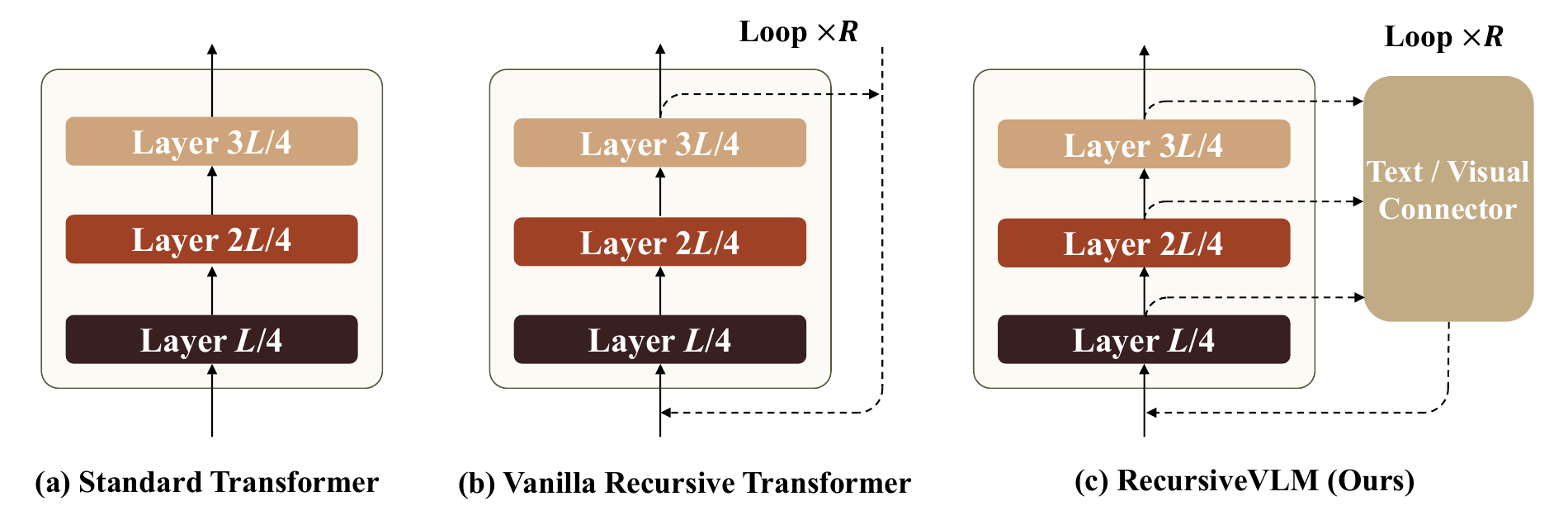}
    \caption{Comparison of recursive architectures for LMMs. (a) Standard non-recursive LMM with single-pass. (b) Vanilla recursion: output embeddings are directly fed back as inputs across steps, causing feature scale and distribution misalignment. (c) Our method fuses features from multiple layers and employs a modality-specific Recursive Connector to bridge recursion steps and mitigate misalignment.}
    \label{fig:taxonomy}
\end{figure*}

\subsection{Large Multimodal Models}
Mainstream LMMs~\cite{gpt4v,gemini,gemini1_5,gemini2_5,qwen,qwen2vl,qwen2_5vl} have achieved great success in vision-language modeling.
A typical LMM architecture consists of three parts: a vision encoder for extracting image features, a large language model serving as the cognitive core, and an adapter module for connecting the vision and language modalities~\cite{llava,minigpt4,blip}.
Recent advances have conducted in-depth exploration in architecture design, such as DeepStack adapter~\cite{deepstack} and modality-independent routers~\cite{ming_omni}. Despite these advances, existing LMMs universally adopt a single forward-propagate paradigm, where each parameter is activated only once per input. To date, there has been no systematic exploration of recursive architectures for LMMs, despite their potential to significantly improve parameter utilization and generalization in multimodal settings. This paper addresses this gap by introducing recursive computation into mainstream LMMs, aiming to enhance both efficiency and adaptability without increasing model size.

\subsection{Recursive Transformers}
Recursive architectures have been introduced to enhance both parameter efficiency and reasoning capabilities in modern Large Language Models (LLMs). 
They can be interpreted through two complementary lenses: parameter sharing and latent reasoning.

\noindent\textbf{Parameter sharing.} From this perspective, recursive models reuse one or more Transformer blocks across depth, effectively sharing weights to reduce model size without increasing computational cost. Early work, such as ALBERT~\cite{albert} and MobileLLM~\cite{mobilellm}, reuses the parameters in the BERT architecture, while the Universal Transformer~\cite{universal_transformers,moeut} shows that a single recurrent block can match the representational capacity of deep non-shared models. Subsequent studies further explore structured sharing strategies: \citet{takase2023lessons} systematically analyzes trade-offs in compression, and Megrez2~\cite{megrez2} reuses MoE experts across layers. However, strict weight tying often leads to representation degradation in large-scale settings. To mitigate this, Relaxed Recursive Transformers~\cite{relaxed_recursive_transformers} introduce lightweight LoRA adapters at each recursion step, while Mixture of Recursions (MoR)~\cite{mor} enables input-adaptive recursion depth by dynamically allocating computation where most beneficial.


\noindent\textbf{Latent reasoning.} Alternatively, recursion can be viewed as iterative refinement of internal representations—a form of latent reasoning~\cite{chen2025inner,geiping2025scaling,saunshi2025reasoning,zeng2025pretraining}. Looped Transformers exemplify this paradigm: they can simulate programmable computation~\cite{looped_programmable}, learn iterative optimization algorithms~\cite{looped_sgd}, generalize to longer sequences on algorithmic tasks~\cite{looped_length}, and model data-fitting dynamics~\cite{looped_learning}. Recent work further validates their scaling potential: \citet{geiping2025scaling} leverages recurrence to scale test-time computation in latent space, and \citet{saunshi2025reasoning} shows that looped models can match much deeper non-recursive counterparts on complex reasoning benchmarks


While these advances highlight the promise of recursive Transformers in LLMs, they largely overlook the unique challenges posed by multimodal settings, particularly severe cross-step feature misalignment and the need for flexible, step-wise predictability. Our work bridges this gap by introducing the \textbf{Recursive Connector} and \textbf{Monotonic Recursion Loss}, enabling stable and adaptive recursion in LMMs for the first time.



%% file: sec/3_method.tex
\section{Method}
\label{sec:method}

\begin{figure*}[t]
\centering
    \begin{subfigure}{0.65\linewidth}
    \centering
    \includegraphics[width=1.0\linewidth]{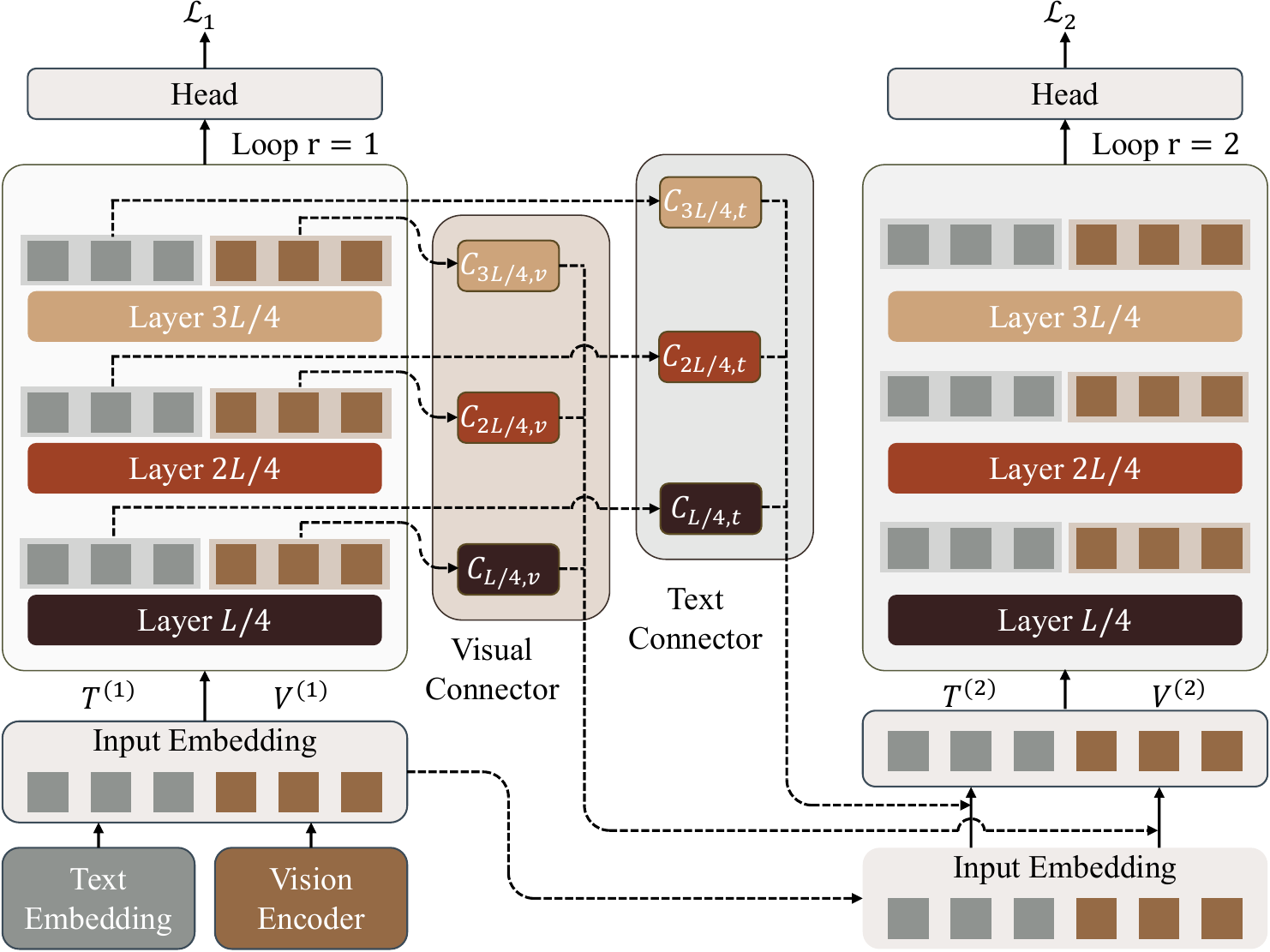}
    \caption{RecursiveVLM }
    \end{subfigure}
\hspace{20pt}
\begin{subfigure}{0.235\linewidth}
    \centering
    \includegraphics[width=1.0\linewidth]{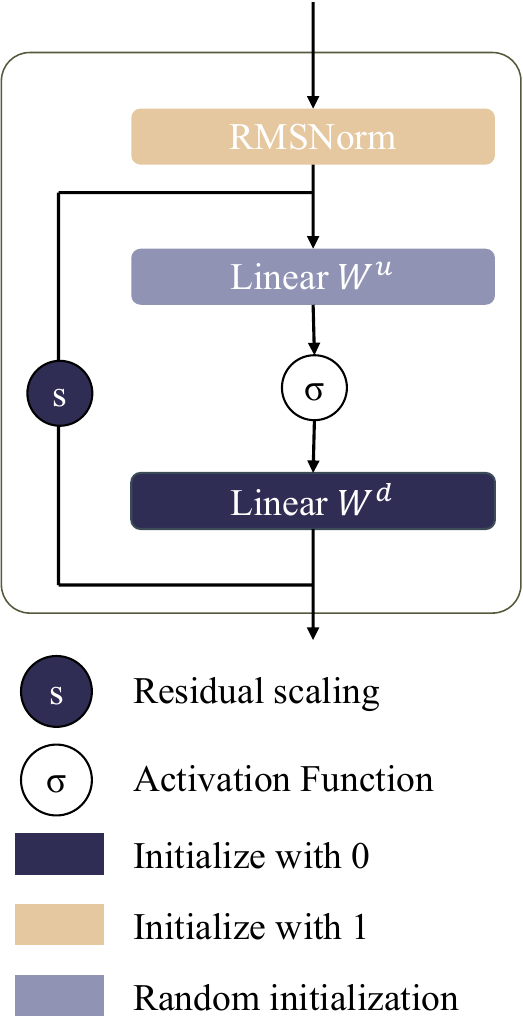}
    \caption{Connector module}
\end{subfigure}
\caption{The framework of our proposed method. (a) RecursiveVLM loops the LLM backbone with a modality-specific multi-layer connector. (b) Detailed structure of the connector module comprising an RMSNorm, a learnable residual scaling vector, and an MLP.}
\label{fig:method}
\end{figure*}

In this section, we present our approach for recasting a vanilla large multi-modal model into a recursive model. We begin with a concise overview of the standard vision-language model framework. Next, we elaborate on the key modifications required to endow the model with recursive processing capabilities.

\subsection{Preliminary}
We start by formalizing a general Vision-Language Model (VLM) architecture.
Given a multimodal input comprising an image and associated text, the vision encoder maps the visual input to a sequence of visual token embeddings, denoted as $\boldsymbol{V}$, and the textual input is encoded into language token embeddings $\boldsymbol{T}$.
These two embedding sequences are then concatenated to form the model’s initial input representation: $\boldsymbol{E} = \left[\boldsymbol{V}, \boldsymbol{T}\right]$.
Prevailing VLMs commonly adopt the Transformer decoder architecture.
Let $\boldsymbol{H}_{\ell}$ denote the hidden state output by the $\ell$-th decode layer, where $\ell=1,2,\cdots,L$, and $L$ is the total number of layers in the decoder.
The transformation at each layer is given by:
\begin{equation}
    \boldsymbol{H}_{\ell} = \mathcal{F}\left(\boldsymbol{H}_{\ell-1}; \boldsymbol{\theta}_{\ell}\right),
\end{equation}
where $\mathcal{F}$ denotes the layer-wise transformation, and $\boldsymbol{\theta}_{\ell}$ represents the trainable parameters associated with the $\ell$-th layer.
The initial hidden state is set to the input embeddings, \emph{i.e.} $\boldsymbol{H}_{0} = \boldsymbol{E}$.
The final output $\boldsymbol{O}$ is derived by decoding the hidden states from the deepest layer $\boldsymbol{H}_{L}$: 
\begin{equation}
    \boldsymbol{O} = \mathcal{G}\left(\boldsymbol{H}_{L};\boldsymbol{\theta}_{\text{head}}\right),
\end{equation}
where $\mathcal{G}$ denotes the final decoding function, and $\boldsymbol{\theta}_{\text{head}}$ represents the parameters of the decoder head.

\subsection{Recursive Transformer}
In this paper, we propose a method that recasts the standard transformer architecture to a recursive one.
Let $r$ denote the index of the recursion step, where $r=1,2,\cdots, R$ and $R$ is the maximum recursion depth.
At the $r$-th recursion step, the model input is denoted as $\boldsymbol{E}^{(r)}$, with the initial input given by $\boldsymbol{E}^{(1)} = \boldsymbol{E}$.
Let $\boldsymbol{H}_{\ell}^{(r)}$ represent the hidden state of the $\ell$-th layer after the $r$-th recursion step.
The layer-wise transformation is defined as:
\begin{equation}
    \boldsymbol{H}_{\ell}^{(r)} = \mathcal{F}\left(\boldsymbol{H}_{\ell-1}^{(r)}; \boldsymbol{\theta}_{\ell}\right),
\end{equation}
where the transformer decoder shares the same set of parameters $\boldsymbol{\theta}_{\ell}$ across all recursion steps.

The key challenge lies in constructing the next-step input $\boldsymbol{E}^{(r+1)}$ from all hidden states of the last recursion $\left\{\boldsymbol{H}^{(r)}_{\ell} \vert \ell\in\left\{1,2,\cdots,L\right\}\right\}$ such that iterative refinement is stable and effective.
Naive recursive methods reuse deep-layer outputs $\boldsymbol{E}^{(r+1)} = \boldsymbol{H}^{(r)}_{L}$, leading to severe feature misalignment as shown in Fig.~\ref{fig:misalignment}.
They also employ a fixed-depth recursion, which limits deployment flexibility.
To address these issues, we introduce two synergistic components:
(i) a Recursive Connector that aligns cross-step representations via multi-layer fusion and modality-aware projection, and
(ii) a Monotonic Recursion Loss that enforces consistent performance improvement across recursion depths.

\subsection{Recursive Connector}

Directly applying $\boldsymbol{E}^{(r+1)}=\boldsymbol{H}^{(r)}_{L}$ suffers from severe \textbf{feature distribution misalignment}, as deep-layer representations diverge significantly from input embeddings in both scale and semantics. As shown in Figure~\ref{fig:misalignment}, the L2 norm and CKA in the LMM indicate severe feature misalignment between hidden states in different layers, thus resulting in performance degradation in the naive recursive architecture.

A modality-specific Recursive Connector is thus proposed to aggregate multi-layer features and map them back to the input embedding space. Specifically, we uniformly sample $k$ intermediate layers from the total $L$ layers, \emph{i.e.} at position $L/4$, $2L/4$, $3L/4$ and $L$ for $k=4$.
For each selected layer $\ell$, we decompose its output into visual and textual components:
\begin{equation}
\boldsymbol{H}_{\ell} = [ \boldsymbol{V}_{\ell}, \boldsymbol{T}_{\ell}],
\end{equation}
where $\boldsymbol{V}_{\ell} \in \mathbb{R}^{N_v \times d}$ and $\boldsymbol{T}_{\ell} \in \mathbb{R}^{N_t \times d}$ denote the hidden states of $N_v$ visual tokens and $N_t$ textual tokens, respectively.
We then apply separate connector modules to each modality, mapping the hidden states back to the input embedding space:
\begin{align}
    \boldsymbol{A}_{\ell,v} =& C_{\ell,v}\left(\boldsymbol{V}_{\ell}\right),\\
    \boldsymbol{A}_{\ell,t} =& C_{\ell,t}\left(\boldsymbol{T}_{\ell}\right),
\end{align}
where each connector $C_{\ell,\ast}$ (shown in Figure~\ref{fig:method}(b)) consists of an RMSNorm layer, a residual connection, and a Multi-Layer Perceptron (MLP). The computation is formulated as:
\begin{align}
    \tilde{\boldsymbol{V}}_{\ell} &= \operatorname{RMSNorm}\left(\boldsymbol{V}_{\ell}\right), \\
    \tilde{\boldsymbol{T}}_{\ell} &= \operatorname{RMSNorm}\left(\boldsymbol{T}_{\ell}\right),
\end{align}
\begin{align}
    \boldsymbol{A}_{\ell,v} &= \tilde{\boldsymbol{V}}_{\ell} \cdot \boldsymbol{s}_{\ell,v} + \sigma\left(\tilde{\boldsymbol{V}}_{\ell} \boldsymbol{W}_{\ell,v}^{u}\right) \boldsymbol{W}_{\ell,v}^{d}, \\
    \boldsymbol{A}_{\ell,t} &= \tilde{\boldsymbol{T}}_{\ell} \cdot \boldsymbol{s}_{\ell,t} + \sigma\left(\tilde{\boldsymbol{T}}_{\ell} \boldsymbol{W}_{\ell,t}^{u}\right) \boldsymbol{W}_{\ell,t}^{d},
\end{align}
where$\operatorname{RMSNorm}(\cdot)$ denotes Root Mean Square Normalization layer~\cite{rmsnorm}. $\boldsymbol{W}_{\ell}^{u}$ and $\boldsymbol{W}_{\ell}^{d}$ are the learnable weight matrices of the MLP’s up-projection and down-projection layers, respectively; $\sigma$ is a nonlinear activation function, and $\boldsymbol{s}_{\ell}$ is a learnable residual scaling vector.

The output of the connector for layer $\ell$ is denoted by $\boldsymbol{A}_{\ell}$.
For brevity, the recursion step index $(r)$ is omitted in this formulation.
$\operatorname{RMSNorm}$ addresses the misalignment in feature scale by normalizing the hidden states.
The MLP is dedicated to resolving the misalignment in feature distribution by mapping the intermediate hidden states into the embedding space, and the residual connection is meant to stabilize training as in previous works~\cite{resnet}.

The input for the subsequent recursion step is constructed in a modality-aware manner. Specifically, let $S=\left\{L/4,\, 2L/4,\, 3L/4,\, L\right\}$ denote the selected layer indices. Visual and textual components are updated separately as:
\begin{align}
    \boldsymbol{V}^{(r+1)} &= \boldsymbol{V}^{(1)} + \sum_{\ell \in S} \boldsymbol{A}_{\ell,v}^{(r)}, \\
    \boldsymbol{T}^{(r+1)}_{t} &= \boldsymbol{T}^{(1)} + \sum_{\ell \in S} \boldsymbol{A}_{\ell,t}^{(r)}.
\end{align}
The full input embedding is then formed by concatenation:
\begin{equation}
    \boldsymbol{E}^{(r+1)} = \left[ \boldsymbol{V}^{(r+1)},\, \boldsymbol{T}^{(r+1)} \right].
\end{equation}
This design fuses semantic cues from multiple layers while preserving the original signal for each modality, enabling more effective parameter utilization across recursion steps.

\noindent\textbf{Initialization Strategy.} To ensure training stability, we initialize the recursive model such that its first recursion step exactly reproduces the behavior of the original pretrained LMM. This is achieved by zero-initializing all connector outputs: (i) the residual scaling vectors $\boldsymbol{s}_{\ell,v}$ and $\boldsymbol{s}_{\ell,t}$ are set to zero, and (ii) following the LoRA initialization scheme, the down-projection weights $\boldsymbol{W}_{\ell,v}^{d}$ and $\boldsymbol{W}_{\ell,t}^{d}$ are also initialized to zero. As a result, at the onset of training, $\boldsymbol{E}^{(r+1)}=\boldsymbol{E}^{(1)}$ for all $r$, guaranteeing that the recursive model reproduces the exact output of the original pretrained model before any adaptation occurs. 



\input{table/main_results}

\subsection{Monotonic Recursion Loss}
To encourage high-quality predictions at every recursion step and avoid performance degradation with increasing recursion depth, we introduce the Monotonic Recursion Loss.

First, for each recursion step $r\in\left\{1, \cdots, R\right\}$, we apply the model’s language modeling head to the final-layer hidden states $H_{L}^{(r)}$ to obtain the prediction for the current step.
We then calculate the standard cross-entropy loss between this prediction and the ground-truth target, which is denoted as $\mathcal{L}_{CE}^{(r)}$.
Second, to enforce consistent performance increase across recursion steps, we impose a per-token monotonicity constraint.
Specifically, we compare the cross-entropy losses of adjacent recursion steps on a token-by-token basis.
For any token $i$ at step $r>1$, if its loss exceeds that of the previous step, which indicates a deterioration in prediction quality, we apply a penalty to its current loss.
Formally, the adjusted per-token loss $\mathcal{L}_{i}^{r}$ is defined as follows: 
\begin{equation}
    \hat{\mathcal{L}}_{i}^{(r)}=
    \left\{
    \begin{aligned}
        &\beta\cdot \mathcal{L}_{i}^{r}, && \text{if } \mathcal{L}_{i}^{(r)}>\mathcal{L}_{i}^{(r-1)},\\
        &\mathcal{L}_{i}^{r},            && \text{otherwise},
    \end{aligned}
    \right.
\end{equation}
where $\mathcal{L}_{i}^{(r-1)}$ denotes the loss for token $i$ at the previous recursion step, and $\beta>1$ is a hyperparameter set as $\beta=1.5$ in our experiments.

The total loss for the $r$-th recursion step is defined as the average of the adjusted per-token losses: $\mathcal{L}^{(r)} = \frac{1}{N}\sum_{i=1}^{N}\hat{\mathcal{L}}_{i}^{(r)}$
where $N$ denotes the number of tokens. For the initial step ($r=1$), we use the standard cross-entropy loss, \textit{i.e.}, $\mathcal{L}_{CE}^{(1)}$. 

The overall training objective is defined as the sum of the losses across all recursion steps:
\begin{equation}~\label{eq:final}
    \mathcal{L}_{\text{total}} = \sum_{r=1}^{R}\mathcal{L}^{(r)}.
\end{equation}
Eq.~\eqref{eq:final} supervises the training to produce accurate prediction at each recursion step. It also explicitly encourages the model to follow a recursive trajectory of monotonic performance enhancement, thereby enabling effective prediction at any intermediate or final recursive step.


%% file: table/main_results.tex
\begin{table*}
    \centering
    \small
    \caption{RecursiveVLM outperforms baselines under both general and CoT training datasets.
    This work is an early attempt on recursive Transformer for LMMs. We thus compare with vanilla recursive baselines as introduced in the LLM domain~\cite {saunshi2025reasoning,ouro}.
    The results show that our method significantly improves performance of standard transfer, thus makes recursive multimodal models practical.
    }
    \setlength{\tabcolsep}{4pt}
    \begin{tabular}{l|cc|cccccccc|c}
    \toprule
    \rowcolor{gray!20}
    & \textbf{Train}  & \textbf{Eval} &  &  &  &  &  &  & \textbf{OCR} & \textbf{Hallusion} &  \\
    \rowcolor{gray!20}
    \multirow{-2}{*}{\textbf{Model}} & \textbf{Steps}  & \textbf{Step} & \multirow{-2}{*}{\textbf{AI2D}} & \multirow{-2}{*}{\textbf{MMStar}} & \multirow{-2}{*}{\textbf{MMVet}} & \multirow{-2}{*}{\textbf{MMMU}} & \multirow{-2}{*}{\textbf{MMB}} & \multirow{-2}{*}{\textbf{MathVista}} & \textbf{Bench} & \textbf{Bench} & \multirow{-2}{*}{\textbf{Avg.}} \\
    \midrule
    \multicolumn{12}{c}{Trained on Data1: general purpose instruction dataset}\\
    \midrule
    Standard Transformer               & 1 & 1 & 69.40 & 45.03 & 55.28 & 40.44 & 67.01 & 51.60 & \textbf{76.20} & 37.56 & 55.31\\
    \midrule
    \multirow{2}{*}{Vanilla recursion~\cite{saunshi2025reasoning,ouro}} & 2 & 1 & 63.02 & 42.38 & 49.82 & 42.67 & 60.65 & 43.27 & 69.90 & 33.55 & 50.66 \\
                                       & 2 & 2 & 62.53 & 41.47 & 51.24 & \textbf{40.89} & 59.36 & 42.10 & 69.00 & 34.61 & 50.15 \\
    \midrule
    Vanilla recursion                  & 2 & 1 & 63.67 & 41.70 & 51.01 & 40.22 & 63.06 & 46.43 & 70.50 & 36.54 & 51.64 \\
     w/ RMSNorm                        & 2 & 2 & 63.18 & 40.28 & 47.80 & 39.11 & 60.48 & 45.40 & 69.40 & 35.15 & 50.10 \\
    \midrule
    \rowcolor{cyan!10}
                                       & 2 & 1 & 72.57 & 51.24 & 59.82 & 38.78 & 73.97 & \textbf{53.83} & 74.70 & 36.41 & 57.46 \\
    \rowcolor{cyan!10}
    \multirow{-2}{*}{RecursiveVLM (Ours)} & 2 & 2 & \textbf{74.64} & \textbf{52.10} & \textbf{59.95} & 40.44 & \textbf{76.63} & 52.17 & 74.70 & \textbf{38.69} & \textbf{58.86} \\
    \midrule
    \multicolumn{12}{c}{Trained on Data2: reasoning-intensive dataset}\\
    \midrule
    Standard Transformer               & 1 & 1 & 66.71 & 50.14 & 63.26 & 38.56 & 65.98 & 55.47 & 82.00 & 46.65 & 58.60\\
    \midrule
    \rowcolor{cyan!10}
                                       & 2 & 1 & 68.78 & 52.16 & 64.04 & 40.00 & 68.56 & 54.83 & 82.90 & 44.90 & 59.52 \\
    \rowcolor{cyan!10}
    \multirow{-2}{*}{RecursiveVLM (Ours)} & 2 & 2 & \textbf{72.57} & \textbf{55.44} & \textbf{65.64} & \textbf{43.44} & \textbf{74.31} & \textbf{58.50} & \textbf{83.70} & \textbf{47.22} & \textbf{62.60} \\
    \bottomrule
    \end{tabular}
    \label{tab:main_results}
\end{table*}

%% file: sec/4_experiments.tex
\section{Experiments}
\label{sec:experiments}

This section first introduces the experimental setup, including the benchmark model, datasets, and evaluation metrics.
It then presents the main performance comparison against baseline methods.
Finally, through a series of ablation studies, we provide an in-depth analysis of the effectiveness of core components of our model, including multi-layer aggregation, layer selection strategy, modality independence, and monotonic recursion loss.

\subsection{Experimental Setup}

\paragraph{Implementation Details.}
All experiments start from the pretrained checkpoint of Ming-Lite-Omni~\cite{ming_omni}, a large omni-modal foundation model with 28 Transformer layers.
We first perform continued pretraining on a hybrid corpus of 6 million multimodal samples and 3 million text-only examples to preserve general language understanding.
Then Supervised Fine-Tuning (SFT) is conducted separately on two distinct instruction datasets:
\begin{itemize}[left=0pt]
    \item \textbf{Data1}: A general-purpose instruction set containing 9 million multimodal question-answering pairs, augmented with 3 million text-only NLP instructions to maintain broad language capability.
    \item \textbf{Data2}: A reasoning-intensive dataset comprising 8 million long Chain-of-Thought (CoT) examples, supplemented with 2.6 million NLP instances to support complex multi-step inference.
\end{itemize}
We use \textbf{Data1} by default unless specified.
For all methods, including the baseline and ablations, we use the same training protocol to ensure the fair comparison. 
During SFT, we set the recursion depth to $R=2$ and apply the Monotonic Recursion Loss with $\beta=1.5$. 
Features are aggregated from layers $[7, 14, 21, 28]$, and models are optimized with the AdamW optimizer using standard hyperparameters. 

\paragraph{Evaluation Benchmarks.}
We selected 8 multimodal evaluation benchmarks to assess model performance, including AI2D~\cite{ai2d}(diagram understanding), MM-Star~\cite{mmstar} (complex multimodal reasoning), MM-Vet~\cite{mmvet} (comprehensive multimodal capability), MMMU~\cite{mmmu} (massive multi-discipline multimodal understanding), MMB~\cite{mmbench} (multimodal understanding capability), MathVista~\cite{mathvista} (mathematical reasoning in visual contexts), OCRBench~\cite{ocrbench} (OCR robustness), and HallusionBench~\cite{hallusionbench} (multimodal hallucination).
These datasets cover a wide range of capabilities from basic perception to complex reasoning, allowing for a comprehensive evaluation of the model's overall performance.

\subsection{Main Results}

To validate the effectiveness of our recursive architecture, we compare it against the standard Transformer~\cite{attention} and vanilla recursive Transformers~\cite{looped_length,saunshi2025reasoning,ouro}.
Results are shown in Table~\ref{tab:main_results}, where the baseline corresponds to our framework with recursion depth $R=1$ \textit{i.e.}, not applying recursion, trained under the same continued pretraining and SFT protocol.
Vanilla recursive methods directly reuse the full output hidden states as input for the next recursion step, a common design in prior work~\cite{looped_length,relaxed_recursive_transformers}.

As shown in Table~\ref{tab:main_results}, our non-recursive baseline consistently outperforms the vanilla recursive methods at both recursion steps with $r=1$ and $r=2$.
The poor performance of vanilla recursive methods can be attributed to the feature misalignment and the lack of modality-aware feature processing, which hinder effective parameter reuse as discussed in Section~\ref{sec:intro}.
A simple remedy, adding RMSNorm to normalize feature scales, fails to resolve these issues and yields negligible gains.
In contrast, our RecursiveVLM achieves consistent improvements across all benchmarks and both recursion steps, demonstrating that carefully designed connectors and monotonic supervision enable stable and effective recursive computation.


\subsection{Ablation Studies}

\subsubsection{The Component of Recursive Connector}
\label{subsec:connector}

\input{table/ablation_connector}

We evaluate the impact of connector components as shown in Table~\ref{tab:ablation_connector}.  
The baseline (\textit{w/o connector}) directly feeds deep-layer features back to the input, causing severe feature misalignment and yielding the worst results.
Incorporating RMSNorm resolved the feature scale misalignment, while still face the challenge of distribution misalignment.
Further incorporating an MLP to project features into the input embedding space yields additional gains on most metrics, though minor fluctuations are observed on MMVet, perhaps due to task-specific sensitivity.
Naively adding a standard residual connection (without learnable scaling) degrades performance on MMStar and MMVet, suggesting that standard identity pathways do not enhance the optimization.
In contrast, our \textbf{scaled residual connection} initializes the scaling vector to zero, which ensures the connector output is zero at the start of training and thus preserves the original Transformer behavior.
It provides a stable initialization and allows the model to gradually learn the optimal balance between the MLP and residual branches.
As a result, the full Recursive Connector achieves the best performance across all benchmarks.


\begin{tcolorbox}[
    enhanced,
    colback=white,  
    colframe=black, 
    width=\linewidth,
    coltitle=white,
    title=\textbf{Takeaway 1},
    top=3mm, bottom=2mm, left=3mm, right=3mm,
    attach boxed title to top left={yshift=-2.8mm, xshift=4mm},
    boxed title style={colback=black, colframe=black, boxrule=0mm, 
    toptitle=1mm, bottomtitle=1mm}
]
Vanilla recursive Transformers are faced with scale and distribution misalignment across recursions.
Our Recursive Connector addresses these issues by normalizing the hidden states with RMSNorm and mapping the hidden states into the input embedding space with an MLP and scaled residual connection, yielding an effective and stable performance improvement.
\end{tcolorbox}

\subsubsection{The effectiveness of Multi-Layer Aggregation}
\label{subsec:multi_layer}

Existing recursive Transformer methods reuse only the final layer output as input for the subsequent recursion step. This design restricts the model’s ability to leverage rich intermediate representations from earlier layers. To investigate the impact of layer selection, we conduct an ablation study on different strategies, with results shown in Table~\ref{tab:ablation_layers}.
Using the last layer alone improves upon the non-recursive baseline, yet may not fully exploit the potential of recursive computation. When selecting two layers, we uniformly sample the middle and final layers (\emph{e.g.}, $[L/2, L]$). This yields moderate gains on some benchmarks but still underutilizes the full hierarchy of hidden states. For four-layer selection, we compare several strategies. Selecting the first four or last four consecutive layers provides an incomplete view of feature evolution across the network, resulting in marginal improvements.
In contrast, our proposed uniform layer selection, sampling layers at regular intervals ($[L/4, 2L/4, 3L/4, L]$), achieves the best performance across all settings. This demonstrates that fusing features from diverse depths, which capture complementary levels of abstraction, is essential for effective recursive refinement.


\input{table/ablation_layers}

\begin{tcolorbox}[
    enhanced,
    colback=white,  
    colframe=black, 
    width=\linewidth,
    coltitle=white,
    title=\textbf{Takeaway 2},
    top=3mm, bottom=2mm, left=3mm, right=3mm,
    attach boxed title to top left={yshift=-2.8mm, xshift=4mm},
    boxed title style={colback=black, colframe=black, boxrule=0mm, 
    toptitle=1mm, bottomtitle=1mm}
]
Vanilla recursive Transformers employ the output of the last layer as the input of the next recursion, which constrains the capability of recursive models. Our Recursive Connector overcomes this bottleneck by aggregating multi-layer features from uniformly spaced depths, enabling richer recursive refinement.
\end{tcolorbox}

\subsubsection{The effectiveness of Modality-Independency}
\label{subsec:modality}

LMMs exhibit distinct feature distributions across vision and language modalities.  
Processing each modality with independent connector parameters allows for decoupled, modality-specific adaptation, thereby enhancing the model’s representational capacity.  
As shown in Table~\ref{tab:ablation_modality}, using a modality-independent connector consistently improves performance over a shared-parameter baseline across most benchmarks, validating the necessity of modality-aware recursive refinement.

\begin{tcolorbox}[
    enhanced,
    colback=white,  
    colframe=black, 
    width=\linewidth,
    coltitle=white,
    title=\textbf{Takeaway 3},
    top=3mm, bottom=2mm, left=3mm, right=3mm,
    attach boxed title to top left={yshift=-2.8mm, xshift=4mm},
    boxed title style={colback=black, colframe=black, boxrule=0mm, 
    toptitle=1mm, bottomtitle=1mm}
]
Vision and language tokens in LMMs follow distinct distributional patterns. Thus modality-independent connectors are essential for recursive adaptation.
\end{tcolorbox}

\input{table/ablation_modality}

\subsubsection{The effectiveness of Monotonic Recursion Loss}
\label{subsec:monotonic_loss}

\input{table/ablation_loss}

Vanilla recursive models are usually equipped with a \emph{Loss at final step}, which is restricted to predict only at a fixed, predefined recursion depth.  
This design cannot generate high-quality outputs at \emph{any} intermediate step, constraining the adaptive and efficient inference across diverse hardware.
As shown in Table~\ref{tab:ablation_loss}, this strategy yields suboptimal performance at the first recursion step (\emph{e.g.}, 68.85\% on AI2D), as no direct signal guides early outputs.
Applying cross-entropy loss at every recursion step (\emph{Loss at each step}) improves $r=1$ performance across all benchmarks, confirming the benefit of multi-step supervision.
However, it does not enforce consistent quality progression, and performance may fluctuate between steps.
In contrast, our \emph{Monotonic Recursion Loss} explicitly penalizes degradation from one step to the next, resulting in the best scores at both $r=1$ and $r=2$.
Notably, it achieves the highest $r=1$ performance, surpassing the per-step loss baseline, while further improving at $r=2$.  
This demonstrates that monotonic supervision not only stabilizes training but also enables high-quality, anytime prediction.


\begin{tcolorbox}[
    enhanced,
    colback=white,  
    colframe=black, 
    width=\linewidth,
    coltitle=white,
    title=\textbf{Takeaway 4},
    top=3mm, bottom=2mm, left=3mm, right=3mm,
    attach boxed title to top left={yshift=-2.8mm, xshift=4mm},
    boxed title style={colback=black, colframe=black, boxrule=0mm, 
    toptitle=1mm, bottomtitle=1mm}
]
Standard recursive training ignores intermediate steps.  
Our Monotonic Recursion Loss enables strong, reliable predictions at \textit{any} recursion step.
\end{tcolorbox}

\subsubsection{The impact of Number of Recursions}
\label{subsec:num_recursions}

The number of recursion steps is a critical hyperparameter in recursive Transformers.  
Prior work has explored varying depths: \citet{relaxed_recursive_transformers} adopts 2-step recursion, while \citet{ouro} reports optimal performance with 4 steps on smaller language models (1.4B–2.6B parameters).
In contrast, we study a large omni-modal model with 16B total parameters and 3B activated per forward pass, hence the optimal recursive step varies.
As shown in Table~\ref{tab:ablation_steps}, 2 recursion steps achieve the best overall performance on standard multimodal benchmarks (\emph{e.g.}, MMStar, MMB).
Further increasing the depth to 3 steps yields marginal gains or even slight degradation on most general tasks, suggesting diminishing returns beyond $R=2$ for these metrics.

However, a striking trend emerges on the HallusionBench: performance consistently improves with deeper recursion (37.56 $\rightarrow$ 38.69 $\rightarrow$ 45.12).
This indicates that recursive refinement helps suppress hallucinatory outputs by iteratively correcting inconsistent or unsupported predictions.
 Notably, our training data and objectives are not tailored for hallucination mitigation; the observed improvement arises purely from the recursive architecture.  
This suggests that recursion inherently promotes more grounded and self-consistent reasoning, a property highly valuable for safety-critical applications.  
We believe this finding opens promising avenues for future work on architecture-driven hallucination reduction.


\input{table/ablation_steps}

\begin{tcolorbox}[
    enhanced,
    colback=white,  
    colframe=black, 
    width=\linewidth,
    coltitle=white,
    title=\textbf{Takeaway 5},
    top=3mm, bottom=2mm, left=3mm, right=3mm,
    attach boxed title to top left={yshift=-2.8mm, xshift=4mm},
    boxed title style={colback=black, colframe=black, boxrule=0mm, 
    toptitle=1mm, bottomtitle=1mm}
]
Recursive multimodal models can refine the predictions during the iterations, which might be promising in suppressing hallucination.
\end{tcolorbox}

%% file: table/ablation_connector.tex
\begin{table}
    \centering
    \small
    \caption{Ablation on Recursive Connector components, highlighting the importance of the scaled residual connection with zero-initialized scaling.
    }
    \setlength{\tabcolsep}{2pt}
    \begin{tabular}{l|c|cccc}
    \toprule
    \rowcolor{gray!20}
    \textbf{Model} & \textbf{Eval Step} & \textbf{AI2D} & \textbf{MMStar} & \textbf{MMVet} & \textbf{MMB} \\ 
    \midrule
    w/o connector & 1 & 62.53 & 41.47 & 51.24 & 59.36 \\ 
    \midrule
    \multirow{2}{*}{RMSNorm}       & 1 & 69.95 & 48.47 & 58.30 & 67.78 \\ 
                  & 2 & 70.79 & 48.91 & 58.62 & 67.87 \\ 
    \midrule
    \multirow{2}{*}{RMSNorm + MLP} & 1 & 70.73 & 48.81 & 57.89 & 70.10 \\ 
                  & 2 & 72.38 & 51.06 & 57.39 & 72.94 \\ 
    \midrule
    RMSNorm + MLP & 1 & 72.60 & 49.82 & 55.09 & 72.59 \\ 
    + residual    & 2 & 73.22 & 49.99 & 57.29 & 73.28 \\ 
    \midrule
    \rowcolor{cyan!10}
    RMSNorm + MLP & 1 & 72.57 & 51.24 & 59.82 & 73.97 \\ 
    \rowcolor{cyan!10}
    + scaled residual & 2 & \textbf{74.64} & \textbf{52.10} & \textbf{59.95} & \textbf{76.63} \\ 
    \bottomrule
    \end{tabular}
    \label{tab:ablation_connector}
\end{table}

%% file: table/ablation_layers.tex
\begin{table}
    \centering
    \small
    \caption{Ablation on the layer selection strategies.}
    \setlength{\tabcolsep}{2pt}
    \begin{tabular}{l|ccccc}
    \toprule
    \rowcolor{gray!20}
    \textbf{Selected layers}      & \textbf{AI2D}  & \textbf{MMStar} & \textbf{MMVet} & \textbf{MMB}   & \textbf{MathVista} \\
    \midrule
    Non-recursive        & 69.40 & 45.03 & 55.28 & 67.01 & 51.60 \\
    $[28]$               & 70.27 & 49.14 & 54.22 & 72.85 & 53.37 \\
    $[14, 28]$           & 72.18 & 48.35 & 56.61 & 70.53 & 52.77 \\
    $[1, 2, 3, 4]$       & 71.15 & 48.30 & 51.83 & 67.53 & 53.27 \\
    $[25, 26, 27, 28]$   & 72.51 & 50.12 & 55.68 & 71.91 & 51.43 \\
    \rowcolor{cyan!10}
    $[7, 14, 21, 28]$ & \textbf{74.64} & \textbf{52.10} & \textbf{59.95} & \textbf{76.63} & \textbf{53.83} \\
    \bottomrule 
    \end{tabular}
    \label{tab:ablation_layers}
\end{table}

%% file: table/ablation_modality.tex
\begin{table}
    \centering
    \small
    \caption{Ablation on modality-shared vs. modality-independent connectors.}
    \setlength{\tabcolsep}{1pt}
    \begin{tabular}{l|ccccc}
    \toprule
    \rowcolor{gray!20}
    \textbf{Connector} & \textbf{AI2D} & \textbf{MMStar} & \textbf{MMVet} & \textbf{MMB} & \textbf{MathVista} \\
    \midrule
    Modality-shared    & 73.19 & 49.80 & \textbf{60.18} & 71.05 & 52.90 \\
    \rowcolor{cyan!10}
    Modality-independent & \textbf{74.64} & \textbf{52.10} & 59.95 & \textbf{76.63} & \textbf{53.83} \\
    \bottomrule
    \end{tabular}
    \label{tab:ablation_modality}
\end{table}

%% file: table/ablation_loss.tex
\begin{table}
    \centering
    \small
    \caption{Ablation on training objectives for recursive models.}
    \setlength{\tabcolsep}{2pt}
    \begin{tabular}{c|c|ccccc}
    \toprule
    \rowcolor{gray!20}
    \textbf{Loss} & \textbf{Eval} &  &  &  &  &  \\
    \rowcolor{gray!20}
    \textbf{Type} & \textbf{Step} & \multirow{-2}{*}{\textbf{AI2D}} & \multirow{-2}{*}{\textbf{MMStar}} & \multirow{-2}{*}{\textbf{MMVet}} & \multirow{-2}{*}{\textbf{MMB}} & \multirow{-2}{*}{\textbf{MathVista}} \\
    \midrule
    Loss at            & 1 & 68.85 & 42.99 & 54.54 & 62.71 & 49.00     \\
    final step         & 2 & 72.02 & 46.82 & 56.56 & 69.07 & 50.50     \\
    \midrule
    Loss at            & 1 & 70.73 & 47.31 & 54.13 & 67.44 & 51.53     \\
    each step          & 2 & 72.57 & 49.71 & 59.73 & 70.19 & 53.67     \\
    \midrule
    \rowcolor{cyan!10}
    Monotonic & 1 & 72.57 & 51.24 & 59.82 & 73.97 & 52.17     \\
    \rowcolor{cyan!10}
    loss      & 2 & \textbf{74.64} & \textbf{52.10} & \textbf{59.95} & \textbf{76.63} & \textbf{53.83}     \\
    \bottomrule
    \end{tabular}
    \label{tab:ablation_loss}
\end{table}

%% file: table/ablation_steps.tex
\begin{table}
    \centering
    \small
    \caption{Ablation on the number of recursion steps.}
    \setlength{\tabcolsep}{2.5pt}
    \begin{tabular}{c|cccccc}
    \toprule
    \rowcolor{gray!20}
    \textbf{Num} &  &  &  &  & \textbf{Hallusion} \\  
    \rowcolor{gray!20}
    \textbf{Recursion} & \multirow{-2}{*}{\textbf{MMStar}} & \multirow{-2}{*}{\textbf{MMVet}} & \multirow{-2}{*}{\textbf{MMB}} & \multirow{-2}{*}{\textbf{MathVista}} & \textbf{Bench} \\  
    \midrule
    1 & 45.03 & 55.28 & 67.01 & 51.60 & 37.56 \\  
    2 & \textbf{52.10} & \textbf{59.95} & \textbf{76.63} & 53.83 & 38.69 \\  
    3 & 51.07 & 55.14 & \textbf{76.63} & \textbf{54.07} & \textbf{45.12} \\  
    \bottomrule
    \end{tabular}
    \label{tab:ablation_steps}
\end{table}

%% file: sec/5_conclusion.tex
\section{Conclusion}
\label{sec:conclusion}
Recursive computation offers a promising path toward parameter-efficient and adaptive inference in LMM, yet existing approaches suffer from feature misalignment and unstable multi-step prediction. To overcome these challenges, we propose RecursiveVLM, a novel recursive Transformer architecture featuring two key innovations: a Recursive Connector that fuses multi-layer, modality-aware representations to enable stable cross-step refinement, and Monotonic Recursion Loss that guarantees non-degrading performance across recursion depths, thereby supporting high-quality prediction at any step.  Notably, our model delivers strong results with minimal computation while enabling progressive refinement when more resources are available, all without increasing model size. This work establishes recursion as an effective paradigm for deployment-adaptive, trustworthy multimodal reasoning, and we hope it inspires further research into structured reuse of model capacity for efficient and controllable AI.


%% file: sec/X_suppl.tex
\clearpage
\setcounter{page}{1}
\maketitlesupplementary
\setcounter{section}{0}

\section{Examples for the refinement of recursive models}
\label{sec:examples}

\begin{table*}[!b]
\begin{tcolorbox}[
    enhanced,
    title=\textbf{Example 1: Object recognition},
    width=\linewidth,
    colback=white,  
    colframe=black,
    coltitle=white,
    top=3mm, bottom=2mm, left=3mm, right=3mm,
    attach boxed title to top left={yshift=-2.8mm, xshift=4mm},
    boxed title style={colback=black, colframe=black, boxrule=0mm,
    toptitle=1mm, bottomtitle=1mm}
]
\centering
\begin{tabular}{p{1.5cm} p{14.0cm}}
     & \includegraphics[width=0.5\linewidth]{fig/mmvet/v1_36.png} \\
\midrule
Question & What should we add in the third step? \\
\midrule
Step 1 & The third step in the pancake recipe involves adding \sethlcolor{red!30}\hl{the dry ingredients to the wet ingredients}. This is typically done by mixing the flour, sugar, baking powder, and salt in a bowl. The image shows a whisk and a bowl, which are used for this purpose. Therefore, the correct answer is:\textbackslash n\textbackslash nB. Flour, sugar, baking powder, and salt. \\
\midrule
Step 2 & The third step in the pancake recipe involves adding \sethlcolor{green!30}\hl{milk to the mixture}. This is indicated by the image showing a bowl with a whisk and a milk carton, suggesting that milk is being poured into the bowl. Therefore, the correct answer is:\textbackslash n\textbackslash nB. Milk. \\
\end{tabular}
\end{tcolorbox}
\caption{RecursiveVLM corrects the critical tokens in the second recursion step, which leads to the right final predictions.}
\end{table*}

In this supplementary material, we present a series of prediction examples of the recursive model.
The RecursiveVLM exhibits improved performance when comparing the predictions of step $1$ and step $2$.
As shown in the following examples, with only $1$ step, \emph{i.e.} vanilla straight-forward Transformers, the wrong prediction tokens can mislead the complete answers.
While with the $2$-step recursion, the model could refine the predictions and correct critical tokens.
This property is essential to multiple vision-language understanding tasks, including object recognition, spatial understanding, mathematical understanding, chart understanding, and common knowledge reasoning.

\begin{table*}[b]
\begin{tcolorbox}[
    enhanced,
    title=\textbf{Example 2: Spatial understanding},
    width=\linewidth,
    colback=white,  
    colframe=black, 
    coltitle=white,
    top=3mm, bottom=2mm, left=3mm, right=3mm,
    attach boxed title to top left={yshift=-2.8mm, xshift=4mm},
    boxed title style={colback=black, colframe=black, boxrule=0mm, 
    toptitle=1mm, bottomtitle=1mm}
]
\centering
\begin{tabular}{p{1.5cm} p{14.0cm}}
     & \includegraphics[width=0.7\linewidth]{fig/mmvet/v1_50.png} \\
\midrule
Question & Which room is bigger, the double garage or the living room? \\
\midrule
Step 1 & \sethlcolor{red!30}\hl{The living room is bigger}.\textbackslash nFinal Answer: Living Room. \\
\midrule
Step 2 & \sethlcolor{green!30}\hl{The double garage has dimensions of 19 feet by 21.25 feet, which gives a total area of 403.75 square feet. The living room has dimensions of 21.5 feet by 18.5 feet, which gives a total area of 397.75 square feet. Therefore, the double garage is bigger}. Final Answer: The double garage. \\
\end{tabular}
\end{tcolorbox}
\caption{The non-recursive prediction ignores the thinking and presents wrong answers. The recursive predictions precisely recognize the rooms and calculate their sizes, leading to the correct comparison.}
\end{table*}

\begin{table*}
\begin{tcolorbox}[
    enhanced,
    title=\textbf{Example 3: Math understanding},
    width=\linewidth,
    colback=white,  
    colframe=black, 
    coltitle=white,
    top=3mm, bottom=2mm, left=3mm, right=3mm,
    attach boxed title to top left={yshift=-2.8mm, xshift=4mm},
    boxed title style={colback=black, colframe=black, boxrule=0mm, 
    toptitle=1mm, bottomtitle=1mm}
]
\centering
\begin{tabular}{p{1.5cm} p{14.0cm}}
     & \includegraphics[width=0.4\linewidth]{fig/mmvet/v1_1.png} \\
\midrule
Question & What is d in the last equation? \\
\midrule
Step 1 & The target equation to solve is 7d-4=11d-9. It is a subtraction problem.The variable in the equation is d. The equation can be equivalent to: \sethlcolor{red!30}\hl{d = (11 - 7) / (7 - 4)}. Then we can get 4. Final answer: 4 \\
\midrule
Step 2 & The target equation to solve is 7d-4=11d-9. It is a subtraction problem.The variable in the equation is d. The equation can be equivalent to: \sethlcolor{green!30}\hl{7d - 11d = -9 + 4}. Then we can get -4d = -5. Finally, the value of d is 5/4. Final answer: 1.25 \\
\end{tabular}
\end{tcolorbox}
\caption{Both steps recognize the correct equations to solve $d$. However, the non-recursive model performs a wrong transformation of the equation. The recursive model is correct.}
\end{table*}


\begin{table*}
\begin{tcolorbox}[
    enhanced,
    title=\textbf{Example 4: Common Knowledge and Reasoning},
    width=\linewidth,
    colback=white,  
    colframe=black,
    coltitle=white,
    top=3mm, bottom=2mm, left=3mm, right=3mm,
    attach boxed title to top left={yshift=-2.8mm, xshift=4mm},
    boxed title style={colback=black, colframe=black, boxrule=0mm, 
    toptitle=1mm, bottomtitle=1mm}
]
\centering
\begin{tabular}{p{1.5cm} p{14.0cm}}
     & \includegraphics[width=0.45\linewidth]{fig/mmvet/v1_151.png} \\
\midrule
Question & Can you explain this meme? \\
\midrule
Step 1 & The meme humorously illustrates the common experience of students in exams, where \sethlcolor{red!30}\hl{they tend to write more in the first and last pages of the exam paper, and less in the middle pages}. \textbackslash n\textbackslash n1. **Top Section - ``My Handwriting In Exams''**: The meme starts with a title that sets the context for the viewer, indicating that the content is about a student's handwriting during exams.\textbackslash n\textbackslash n2. **First Two Pages**: The text here reads, ``Global warming is definitely a very huge issue in front of this entire world. It's causing several disturbances in the.'' This suggests that the student has written a significant amount of text in the first two pages, possibly about the topic of global warming.\textbackslash n\textbackslash n3. **Middle Pages**: The text here reads, \sethlcolor{red!30}\hl{``Also it can cause the melting of glaciers and melting of sea ice, which can make the sea level rise.''} This indicates that the student has written about the consequences of global warming, such as the melting of glaciers and sea ice, and the resulting rise in sea levels.\textbackslash n\textbackslash n4. **Last Two Pages**: The text here reads, \sethlcolor{red!30}\hl{``The last two pages are about the impact of global warming on the environment and the economy.''} This suggests that the student has written about the broader impacts of global warming, including its effects on the environment and the economy.\textbackslash n\textbackslash nThe meme uses a simple and relatable scenario to highlight a common experience, making it humorous and easy to understand. \\
\midrule
Step 2 & The meme humorously illustrates the phenomenon of students \sethlcolor{green!30}\hl{starting strong in exams but losing focus and energy as they progress}. The top section, labeled ``First two pages,'' shows a neat and focused handwritten text about global warming. The middle section, ``Middle pages,'' shows a decline in handwriting quality, with the text becoming less coherent and more disorganized. The bottom section, ``Last two pages,'' shows a significant drop in handwriting quality, with illegible scribbles and a lack of coherence, symbolizing the student's struggle to maintain focus and energy towards the end of the exam. The image is a light-hearted take on the common experience of students feeling tired or distracted during exams. \\
\end{tabular}
\end{tcolorbox}
\caption{An example illustrates how a recursive model improves on common knowledge and reasoning. The non-recursive model does not understand the meme, reading non-existent words in the image. The recursive model, on the other hand, can understand the memes and correctly recognize the content in the image.}
\end{table*}

\begin{table*}[t]
\begin{tcolorbox}[
    enhanced,
    title=\textbf{Example 5: Chart understanding},
    width=\linewidth,
    colback=white,  
    colframe=black, 
    coltitle=white,
    top=3mm, bottom=2mm, left=3mm, right=3mm,
    attach boxed title to top left={yshift=-2.8mm, xshift=4mm},
    boxed title style={colback=black, colframe=black, boxrule=0mm, 
    toptitle=1mm, bottomtitle=1mm}
]
\centering
\begin{tabular}{p{1.5cm} p{14.0cm}}
     & \includegraphics[width=0.8\linewidth]{fig/mmvet/v1_25.png} \\
\midrule
Question & Based on the image, what is the total international awareness of Tina Fey? \\
\midrule
Step 1 & The image shows the international awareness of Tina Fey for different territories.\textbackslash nThe international awareness \sethlcolor{red!30}\hl{for the UK is 45\%}.\textbackslash nFinal Answer: 45\% \\
\midrule
Step 2 & \sethlcolor{green!30}\hl{The international average for Tina Fey is 35\%}.\textbackslash nFinal Answer: 35\%. \\
\end{tabular}
\end{tcolorbox}
\caption{The non-recursive model encounters hallucination, as it predicts ``for the UK'' which is not mentioned above. The recursive model finds the correct answer directly.}
\end{table*}